\documentclass[authorversion, nonacm]{acmart}

\usepackage{caption}
\usepackage{subcaption}
\AtBeginDocument{%
  }

\setcopyright{acmlicensed}
\copyrightyear{2025}
\acmYear{2025}
\acmDOI{XXXXXXX.XXXXXXX}

\acmConference[CHI '25]{Proceedings of the 2025 CHI Conference on Human Factors in Computing Systems}{Apr 26--May 01,
  2025}{Yokohama, Japan}
\acmISBN{978-1-4503-XXXX-X/18/06}

\begin{document}

\title[Predicting Human Behavior in Autonomous Systems]{Predicting Human Behavior in Autonomous Systems: A Collaborative Machine Teaching Approach for Reducing Transfer of Control Events}



\author{Julian Wolter}
\affiliation{
   \institution{DFKI, Saarland Informatics Campus}
   \city{Saarbr\"ucken}
   \country{Germany}}
\email{julian.wolter@dfki.de}

\author{Amr Gomaa}
\affiliation{
   \institution{DFKI, Saarland Informatics Campus}
   \city{Saarbr\"ucken}
   \country{Germany}}
\email{amr.gomaa@dfki.de}


\renewcommand{\shortauthors}{J. Wolter et al.  }

\begin{abstract}

As autonomous systems become integral to various industries, effective strategies for fault handling are essential to ensure reliability and efficiency. Transfer of Control (ToC), a traditional approach for interrupting automated processes during faults, is often triggered unnecessarily in non-critical situations. To address this, we propose a data-driven method that uses human interaction data to train AI models capable of preemptively identifying and addressing issues or assisting users in resolution. Using an interactive tool simulating an industrial vacuum cleaner, we collected data and developed an LSTM-based model to predict user behavior. Our findings reveal that even data from non-experts can effectively train models to reduce unnecessary ToC events, enhancing the system’s robustness. This approach highlights the potential of AI to learn directly from human problem-solving behaviors, complementing sensor data to improve industrial automation and human-AI collaboration.

\end{abstract}


\begin{teaserfigure}
  \centering
  \includegraphics[width=\textwidth]{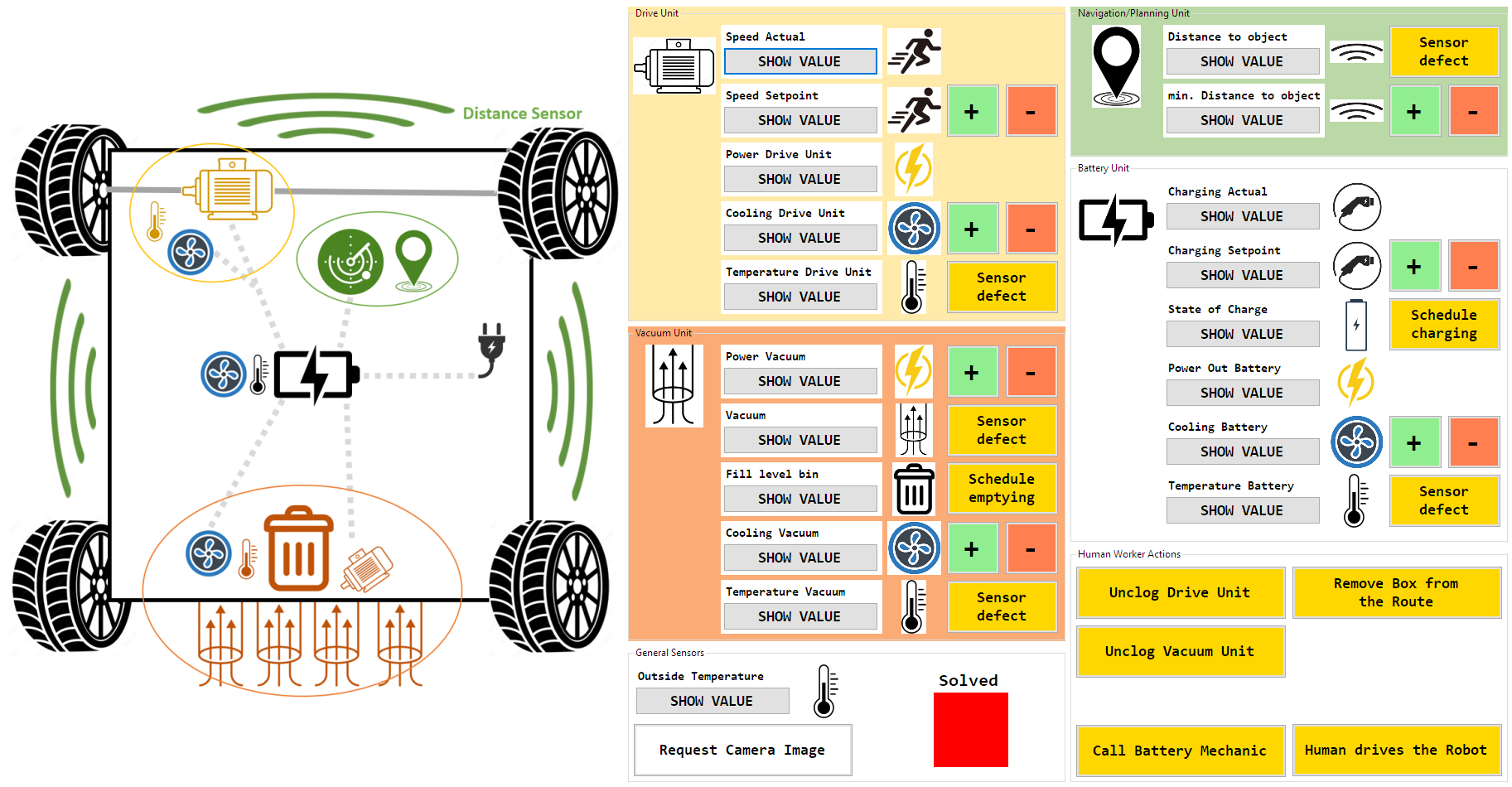}
  \caption{Our Interactive Data Collection Tool Used to Gather Annotated Data for Training Our Behavior Prediction LSTM Model.}
  \Description{Our Interactive Data Collection Tool Used to Gather Annotated Data for Training Our Behavior Prediction LSTM Model.}
  \label{fig:teaser}
\end{teaserfigure}


\maketitle

\section{Introduction}

Autonomous systems are becoming increasingly prevalent in various industries, yet effective troubleshooting and maintenance remain critical. This study explores the potential of using human interaction data from non-experts to train AI models for resolving issues in autonomous systems, specifically in our use case: an industrial vacuum cleaner. The motivation behind this research stems from the concept of Transfer of Control (ToC), which involves the handover of control from an autonomous system to a human operator when the system encounters a problem it cannot resolve on its own.
Previous research has shown that ToC can be a significant bottleneck in the efficiency of autonomous systems, often leading to unnecessary interruptions and delays. Traditional approaches to ToC typically involve predefined rules or thresholds that trigger the transfer. Still, these methods can be rigid and fail to account for the nuances of real-world problem-solving. For instance, studies have highlighted the limitations of rule-based systems in dynamic environments, where unexpected issues can arise that were not anticipated during the system’s design~\cite{goodrich2008human,chen2014situation, kleer2024towards}.

Our approach leverages the sequences of actions taken by operators to train an AI model that can preemptively address issues or assist new operators in resolving robot faults. By learning from human problem-solving behaviors, AI can reduce unnecessary ToC events and improve the overall efficiency and reliability of the system. This method contrasts with previous work by focusing on dynamic, data-driven learning rather than static rules, allowing for more adaptive and context-aware interventions. Recent advancements in machine learning, particularly in the use of Long Short-Term Memory (LSTM) networks for sequence prediction, have shown promise in capturing complex patterns in user interactions~\cite{hochreiter1997long,greff2016lstm,zhao2018human,laplaza2024enhancing}.

In this paper, we highlight the potential for AI to support industrial operations by learning from human problem-solving behaviors. By leveraging previous sequences of actions and knowledge about the robot that is already there, our LSTM-based method can be trained to preemptively address issues or assist users in resolving constraints, thereby enhancing the efficiency and reliability of autonomous systems by reducing the number of interruptions caused by ToC requests. This approach not only improves system performance but also reduces the cognitive load on human operators, allowing them to focus on more complex tasks that require human intuition and expertise.

\section{Background and Related Work}

LSTM networks have gained popularity in modeling sequential data, making them particularly suitable for tasks involving temporal dependencies and sequence predictions. In the context of robotic planning, LSTMs are effective at capturing the sequential nature of human actions \cite{Maithani2019, lin2022predicting} and system states, which is essential for predicting future events based on past behavior or detecting \cite{wang2023fault} and resolving faults. Obtaining the training data for such a model is a major challenge. This often requires interaction with experts, which is difficult to collect, or the data is generated synthetically \cite{ding2020model}, which is not always possible.

When provided with a good dataset, LSTMs have proven to be highly effective in predicting human behavior in robotic systems \cite{how2014multiple}. These models can accurately anticipate human actions based on patterns in the data. Additionally, LSTMs are capable of generating plans for robot execution, allowing robots to autonomously perform tasks. While predicting human behavior and generating robot plans are distinct applications, LSTMs are well-suited for both tasks, enhancing the capabilities of robotic systems in dynamic environments \cite{molina2021trajectory, fong2021model, al2020recurrent}. We have therefore decided to also choose an LSTM for our approach and try to integrate more data.

Integrating knowledge into LSTM networks is challenging due to their inherent design for learning from sequential data without explicit knowledge structures. Several approaches have been developed to address this, e.g. Hybrid Models, which combine LSTMs with rule-based systems \cite{xie2021embedding}. Our approach involves embedding custom features, where domain knowledge is encoded into feature embeddings that the LSTM can use as additional input \cite{Dash_2022}. 






\section{Methodology}

In autonomous systems, "Transfer of Control" (ToC) events often have similar or even identical causes, leading to multiple occurrences under comparable conditions. This pattern suggests that the triggers for these ToCs are frequently consistent across different situations. By recognizing and understanding these recurring causes, the system can be trained to learn from past ToCs, allowing it to better manage or prevent similar events in the future.

Our approach leverages this capability to learn from previous ToCs and incorporates additional information about the robots's structure. Many autonomous systems are equipped with a digital twin or at least a taxonomy that provides detailed insights into how sensors and actuators relate to each other. By utilizing this additional information, we can further enhance the system's ability to support the operator during a ToC and, ideally, prevent such events from happening altogether.

Moreover, our approach emphasizes learning from interactions with non-experts. Users may not always resolve issues directly and might require further diagnostic steps to identify the problem. These sequences of diagnostic interactions are also valuable and should be included in the system's training process to enhance its ability to learn from a wide range of interactions and scenarios.

We have chosen to use an LSTM model along with a taxonomy to represent the knowledge about the robot. 
By training the LSTM on data from past ToC events handled by non-expert users, we can create a system that continually enhances its predictive abilities and reduces the frequency of ToC events.

\subsection{Taxonomy}

\begin{figure}[t]
  \includegraphics[width=0.85\textwidth]{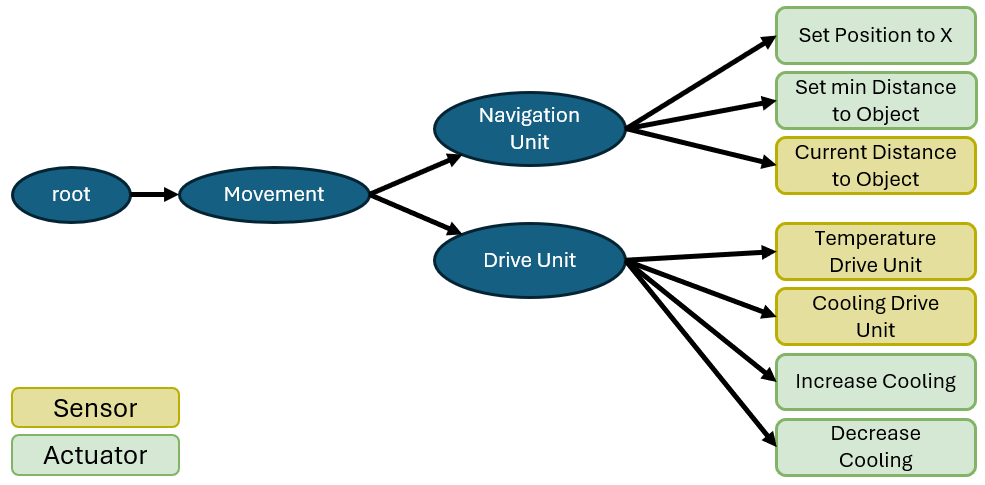}
  \caption{Excerpt From the Taxonomy of the Simulated Industrial Vacuum Robot Used in Our User Study.}
  \Description{Excerpt From the Taxonomy of the Simulated Industrial Vacuum Robot Used in Our User Study.}
  \label{fig:taxonomy}
\end{figure}

The taxonomy used in our system is a structured representation of the domain knowledge for the simulated industrial vacuum robot, organized hierarchically to classify the system's components. It starts with a root node and extends through two layers of intermediate categories, leading to a third layer that contains the specific entities or leaf nodes. These leaf nodes are divided into two primary types: sensors and actuators, representing the core elements of the system.


An excerpt from the taxonomy is shown in Figure \ref{fig:taxonomy}, which illustrates this hierarchical structure and the classification of sensors and actuators within their respective categories.

\subsection{LSTM}

To model and predict user interactions within our system, we utilized an LSTM model. The sequences of user interactions were encoded using tokens, where each token represented a single sensor reading or action performed by the user. To facilitate sequence processing, start and stop tokens were added to each sequence.

\begin{figure}[t]
  \includegraphics[width=0.75\textwidth]{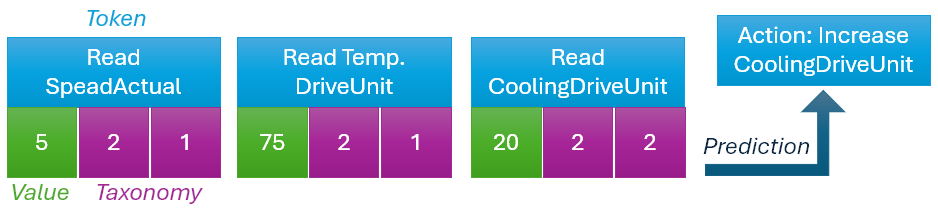}
  \caption{Example of a Diagnosis Sequence for the Error “Robot Driving Slow,” With the Predicted Solution.}
  \Description{Example of a Diagnosis Sequence for the Error “Robot Driving Slow,” With the Predicted Solution.}
  \label{fig:lstm-pipeline}
\end{figure}

The encoding of these sequences involved two key components: sensor values and the taxonomy hierarchy. Sensor values, which are non-discrete numerical data, and the hierarchical structure of the taxonomy were both encoded using embeddings. This approach allowed us to effectively represent the continuous sensor values and the categorical relationships defined in the taxonomy within the LSTM model. An example of this encoding process is provided in Figure \ref{fig:lstm-pipeline}, which illustrates how sensor values and taxonomy hierarchy are translated into embeddings for the LSTM.

Our model architecture was implemented in PyTorch with CUDA acceleration to leverage GPU capabilities for faster computation and training. Given that our model was designed without constraints, it was theoretically possible for the LSTM to predict the same action multiple times consecutively or to predict a "read sensor" operation when an "action" was required. However, this was not an issue in practice since our training data did not contain such data.

The LSTM architecture employed was a many-to-one configuration, focused on predicting the next token in the sequence. Instead of directly predicting the embeddings for sensor values and taxonomy, the LSTM model predicts the next token based on the input sequence. After each prediction, the predicted token is added to the end of the sequence, and the associated embeddings are appended to the new token. This process is repeated until the solution to the problem is found, continuously updating the sequence and predicting the next token based on the updated context.

\section{User Study \& Results}

\subsection{Study}

To gather data for our research, we conducted a user study involving 30 participants interacting with a simulated industrial vacuum robot. This scenario aimed to enable participants to immerse themselves as much as possible in the task, even without extensive experience in an industrial environment. The simulated robot was equipped with 20 read-only sensors and 26 possible actions that participants could trigger. Sensor readings were initially hidden from the participants and could only be revealed through user interaction. Each time a participant clicked to reveal a sensor reading, the action was logged to provide insights into their diagnostic process. To further replicate real-world conditions, we added noise to the sensor values.

To aid participants, we provided a taxonomy-based representation of the robot’s knowledge, visually indicated through colour-coding and grouping of components as you can see in Figure \ref{fig:teaser}. This approach helped participants quickly understand the relationships between sensors and actuators, aiding in diagnosing and resolving issues. Similarly, our system utilized this taxonomy knowledge by creating embeddings that capture these relationships, enhancing the predictive model’s understanding.

The study consisted of two parts: an exploration phase for learning the system and a data collection phase. During the data collection phase, participants were presented with a general error and had to diagnose it by revealing sensor values through clicks. If their initial diagnosis was incorrect, they could continue diagnosing by revealing more sensor values and making additional attempts to identify the issue. Once they identified the problem, they resolved it by triggering the correct actions. Multiple actions could be required, reflecting the complexities of real-world problem-solving.

We removed sequences that were outliers in terms of their length, which occurred 24 times. Additionally, we excluded sequences that did not lead to a successful resolution of the problem, accounting for 6 instances. After applying these filters, we obtained 570 valid sequences for training. The average length of these sequences was 12.8 steps, providing a robust dataset for our model.

We also analyzed the action-to-read ratio, which was 15.3\%. This ratio represents the proportion of actions triggered by participants about the number of sensor readings revealed. In comparison, the reference resolution had an action-to-read ratio of 7.8\%. This lower ratio indicates that fewer sensor readings were necessary per action to resolve an issue in the reference solution. This means that our participants read more sensor values than necessary to resolve the issues. However, this is to be expected, as the participants were non-experts for the system and may have needed additional information to make confident decisions. This broadens the applicability of our system, making it more versatile and practical for real-world deployment where expert data may not always be readily available in contrast to non-expert data.

\begin{figure}[H]
  \includegraphics[width=\textwidth]{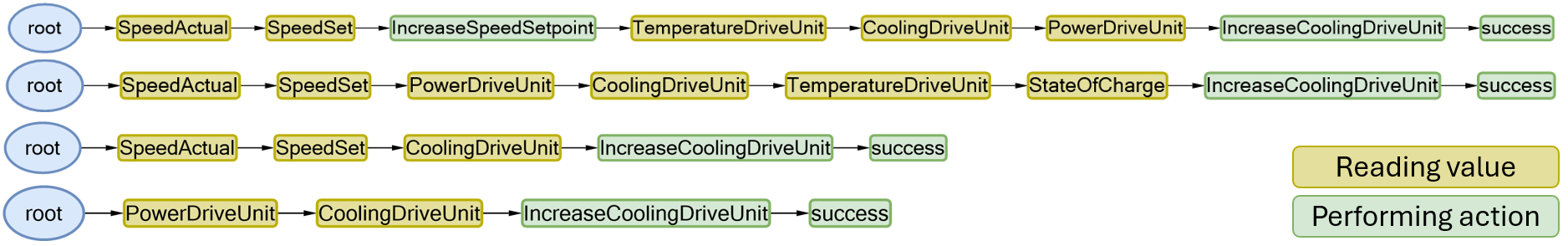}
  \caption{Example Sequences for the Error “There Is Still Dirt Behind the Robot” From the Dataset.}
  \Description{Example Sequences for the Error “There Is Still Dirt Behind the Robot” From the Dataset.}
  \label{fig:exp-dataset}
\end{figure}

An example sequence for the error “There is still dirt behind the robot” is shown in Figure \ref{fig:exp-dataset}, illustrating a typical diagnostic and resolution process used by participants in the study. This example helps to visualize the types of sequences included in our training dataset and provides insights into user behavior and interaction patterns with the system.

After collecting the data, we split it into three subsets: 80\% for training, 10\% for validation, and 10\% for testing and trained our LSTM model using the approach explained.

\subsection{Correctness Metric}

Defining what constitutes a "correct" diagnosis in our system was particularly challenging. Diagnosing complex errors often involves multiple steps and can vary significantly depending on the approach taken by different individuals. For non-experts, it is unrealistic to expect that they will always provide the shortest or quickest diagnostic path. Non-expert users may take longer routes or additional steps in their diagnostic process. However, these detours are not necessarily incorrect or disadvantageous, especially when performed automatically by the system, as they do not consume additional time from human operators. In fact, a more thorough diagnostic process can sometimes be more beneficial than stumbling upon the correct solution by chance. A detailed diagnosis may provide deeper insights into the system's behavior, uncovering underlying issues that might otherwise go unnoticed. Therefore, our correctness metric should account for these variations in diagnostic paths.

Therefore, we decided to use two correctness metrics for our evaluation:
\begin{itemize}
    \item \textbf{Prediction Correctness}: A prediction is considered correct if the predicted step occurs at any point in the test sequence.
    \item \textbf{Sequence Correctness}: A sequence is considered correct if it successfully resolves the problem by executing the correct action or actions.
\end{itemize}

By using these two metrics, we can effectively evaluate the usefulness of individual predictions as well as the overall success of the predicted sequences in solving the problem.

\subsection{Evaluation}

\begin{figure}[h]
\centering
\begin{subfigure}{.55\textwidth}
  \centering
  \includegraphics[width=\textwidth]{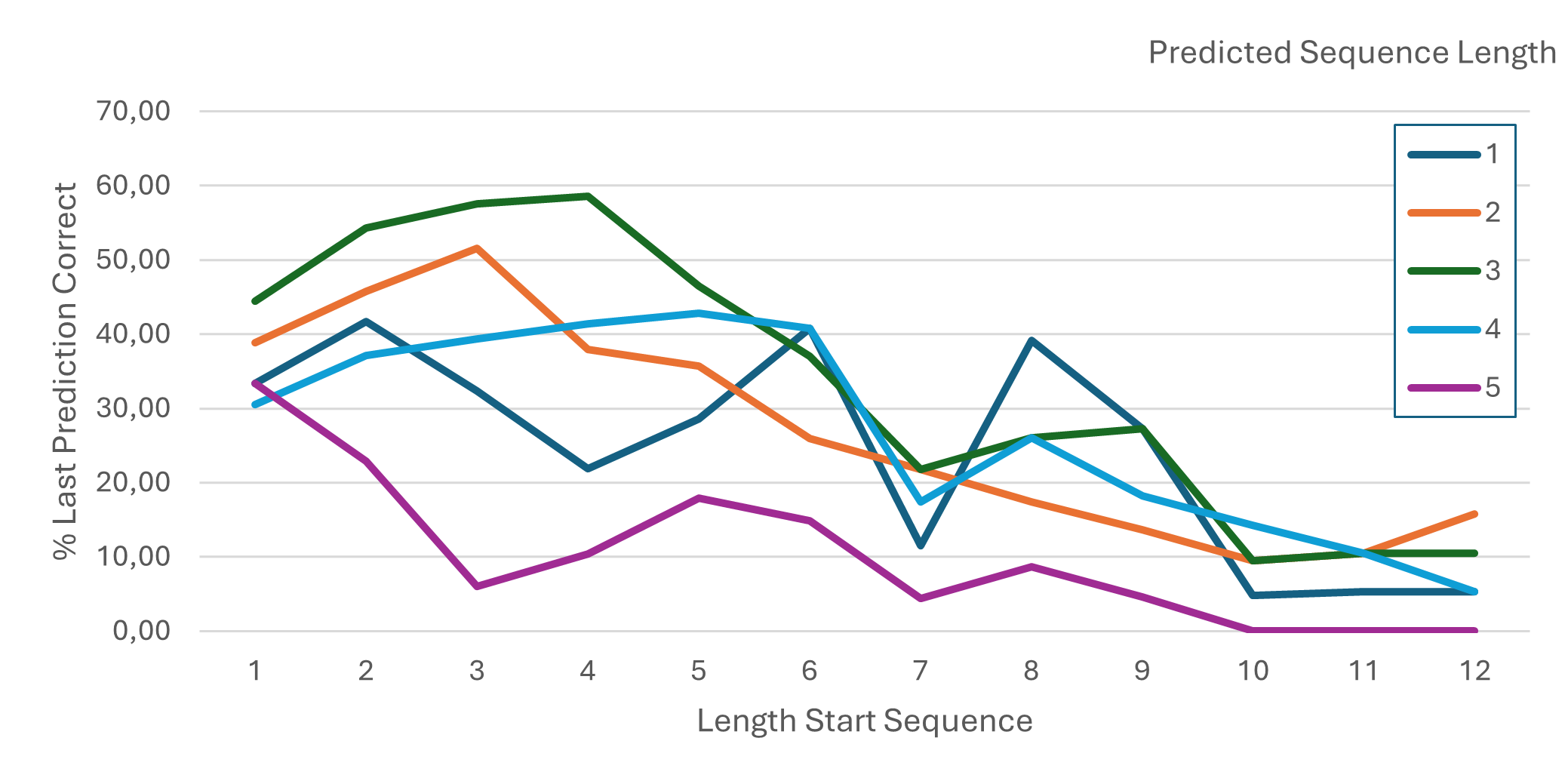}
  \caption{Predicting 1-5 Steps From a Variable Start Length and Check if Last Prediction Is Correct.}
  \Description{Predicting 1-5 Steps From a Variable Start Length and Check if Last Prediction Is Correct.}
  \label{fig:evaluation-2}
\end{subfigure}%
\hspace{.03\textwidth}
\begin{subfigure}{.35\textwidth}
  \centering
  \includegraphics[width=\textwidth]{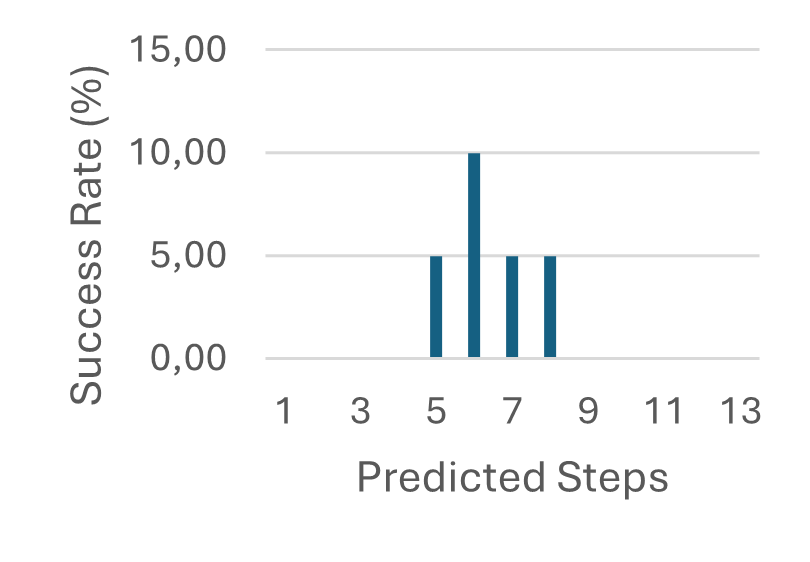}
  \caption{Successful Solutions With Empty Start Sequence Until the First Action.}
  \Description{Successful Solutions With Empty Start Sequence Until the First Action.}
  \label{fig:evaluation-3}
\end{subfigure}
\caption{Results for predicting the user behaviour using LSTMs.}
\end{figure}

In an initial evaluation, we used start sequences of variable lengths and tasked the model with predicting the next 1-5 steps. Following this, we assessed the predictions based on the previously defined correctness metrics to determine if the final predicted step in each sequence was correct. The results of this evaluation are shown in Figure \ref{fig:evaluation-2}.

As we can observe, the prediction accuracy drops significantly after the 4th prediction. Additionally, predictions made with shorter start sequences are noticeably more accurate than those with longer ones. This difference can be explained by the fact that the participants in the study were non-experts. Consequently, the sequences often included unnecessary steps or detours, meaning that longer start sequences already contained non-essential diagnostic actions that could confuse the system. We also noticed that when a prediction was incorrect, the system was able to recover in the subsequent step. This demonstrates the robustness of our approach, which is designed to handle data from non-experts effectively. 


In another evaluation, we simulated all possible errors and started the system with an empty start sequence. The system autonomously performed all steps sequentially until it executed an action. We then checked whether the action resolved the problem and counted the number of steps taken to reach the solution. The results of this evaluation are shown in Figure \ref{fig:evaluation-3}.

Based on the 20 problems presented in the study, our approach was able to autonomously solve five of them resulting in a 25\% success rate, which is significantly better than random chance (around 3\%). In these cases, the number of steps needed to reach a solution was often only slightly more than those required by the ideal solution. This further demonstrates that using data from non-experts for training can still lead to meaningful and effective results.


\section{Conclusion}

We have developed an approach to handle ToC events by predicting human behavior. To achieve this, we created a simulated industrial vacuum robot with a complex error model and engaged non-expert users to solve specific problems with it, collecting data from their interactions. This data, combined with a taxonomy containing domain-specific knowledge, formed the dataset used to train an LSTM model. This model serves two primary purposes: it can act as a support tool that builds upon existing sequences to make predictions, aiding the operator when they are unsure of the next step, or it can autonomously resolve a ToC event without human input. Our results show performance significantly better than random chance, demonstrating the potential of using non-expert data to develop robust systems that effectively manage ToC events, thereby reducing the need for human intervention in autonomous systems.


\begin{acks}
   This work is supported by the German Federal Ministry of Education and Research in the project CAMELOT (grant no. 01IW20008).
\end{acks}

\bibliographystyle{ACM-Reference-Format}
\bibliography{sample-base}


\end{document}